\documentclass[journal]{IEEEtran}
\usepackage{style}

\begin{document}

\title{Diffusion Model for Planning:\\A Systematic Literature Review}

\author{\IEEEauthorblockN{Toshihide Ubukata\IEEEauthorrefmark{1},
Jialong Li\IEEEauthorrefmark{1,3}and
Kenji Tei\IEEEauthorrefmark{2}}

\IEEEauthorblockA{\IEEEauthorrefmark{1} Waseda University, Tokyo, Japan}\\
\IEEEauthorblockA{\IEEEauthorrefmark{2} Tokyo Institute of Technology, Tokyo, Japan}\\
\IEEEauthorblockA{\IEEEauthorrefmark{3} Corresponding Author: lijialong@fuji.waseda.jp}
}


\markboth{}%
{}

\maketitle
\begin{abstract}
Diffusion models, which leverage stochastic processes to capture complex data distributions effectively, have shown their performance as generative models, achieving notable success in image-related tasks through iterative denoising processes.
Recently, diffusion models have been further applied and show their strong abilities in planning tasks, leading to a significant growth in related publications since 2023.
To help researchers better understand the field and promote the development of the field, we conduct a systematic literature review of recent advancements in the application of diffusion models for planning.
Specifically, this paper categorizes and discusses the current literature from the following perspectives: (i) relevant datasets and benchmarks used for evaluating diffusion model-based planning; (ii) fundamental studies that address aspects such as sampling efficiency; (iii) skill-centric and condition-guided planning for enhancing adaptability; (iv) safety and uncertainty managing mechanism for enhancing safety and robustness; and (v) domain-specific application such as autonomous driving.
Finally, given the above literature review, we further discuss the challenges and future directions in this field.
\end{abstract}

\begin{IEEEkeywords}
Diffusion Model, Planning, DDPM, Planner
\end{IEEEkeywords}
\IEEEpeerreviewmaketitle

\section{Introduction} \label{sec:intro}
In the field of Generative Artificial intelligence (GenAI) models, Diffusion models\cite{sohldickstein2015deep, ho2020denoising} have recently emerged as a powerful class of generative models. These models utilize stochastic processes to transform random noise into high-quality data through iterative denoising. Diffusion models have shown great capabilities in image-related tasks, such as generation \cite{zhang2023survey, moser2024diffusion, zhang2023texttoimage}, restoration and enhancement \cite{li2023diffusion}, and editing \cite{huang2024diffusion}.

The fundamental principle of diffusion models involves introducing noise to training data and learning to reverse this process through iterative denoising, effectively capturing complex data distributions. Technically, diffusion models use stochastic differential equations to progressively refine noisy inputs into coherent outputs, enabling accurate modeling of the underlying data distribution. This iterative refinement process allows diffusion models to explore a wide solution space, generating high-quality, diverse outputs.

Thanks to the above specific characteristics of diffusion models, their application extends beyond image processing and has recently been applied in planning tasks, especially in high-dimensional scenarios like motion planning. Recent studies have increasingly reported advantages of diffusion models, such as robustness, highlighting their potential as planners. This trend is also evidenced by the increasing number of related publications, particularly since 2023, as shown in Figure \ref{fig:chart_paper_num}.
In this context, we believe that planning based on diffusion models is a promising research area. However, due to their technical diversity and wide applicability, comprehensively understanding diffusion model-based planning is a challenge.

\begin{figure}[h!]
    \centering
    \includegraphics[width=0.95\linewidth]{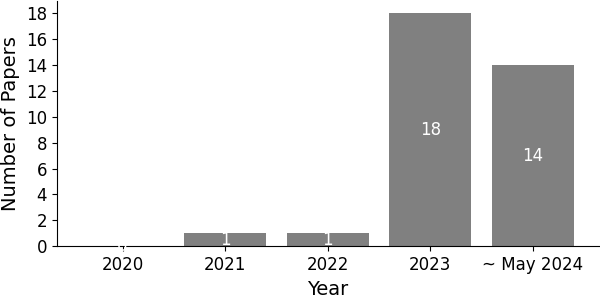}
    \caption{Trends in paper numbers searched using keywords ``Diffusion" and ``Plan" in ArXiv.} 
    \label{fig:chart_paper_num}
\end{figure}

To this end, we conduct a Systematic Literature Review (SLR) to comprehensively examine the applications of diffusion models in planning. Specifically, we reviewed 41 relevant and filtered papers and categorized them based on their research focus (e.g., safety, efficiency). Additionally, we compared the relationships and differences in ideas within the same category of papers. We hope this paper can provide a comprehensive overview, making it easier for researchers and practitioners to understand the state of the art of diffusion models in the field of planning.

The structure of this paper is as follows: 
\hyperref[sec:search_method]{Section II} details the methodology used for the searching and filtering literature.
\hyperref[sec:dataset]{Section III} delves into the various datasets employed to evaluate diffusion models, categorizing them based on specific tasks such as motion planning, path planning, and reinforcement learning.
\hyperref[sec:general]{Section IV} covers the foundation of the diffusion model in planning, motion and path planning, efficiency improvements, preference customization, and hierarchical structures against traditional planning methods.
\hyperref[sec:skillcond]{Section V} explores skill-centric and condition-guided planning approaches.
\hyperref[sec:rob]{Section VI} addresses the enhancement of safety and robustness in planning, including safety mechanisms and strategies for managing uncertainty.
\hyperref[sec:app]{Section VII} discusses specific applications, such as 3D planning and optimization, instructional video procedure planning, autonomous driving, and multi-task diffusion models.
After reviewing all the preceding sections,  \hyperref[sec:challenges]{Section VIII} discusses future challenges, and \hyperref[sec:conclusion]{Section IX} concludes the paper by summarizing the key points discussed.

\section{Literature Search Methodology} \label{sec:search_method}
To ensure a comprehensive survey of the current state of the art in diffusion models for planning, we employed a literature search strategy.

We conducted our literature survey on IEEE Xplore, ACM Digital Library, Dblp, and ArXiv. We focused on papers published from 2020, following the release of the first paper on Diffusion Models\cite{ho2020denoising}, until May 31, 2024.
The primary keywords used in our search were:
\begin{itemize}
    \item Diffusion: ``diffusion'', ``diffuser'', "DDPM"
    \item Planning: ``plan'', ``replan''  + \{``-er'', ``-ing''\} 
\end{itemize}

Our search was specifically targeted at papers related to diffusion-based planning in the reinforcement learning diagram, explicitly excluding those that focused on large language models (LLMs) or natural language processing (NLP).
Initially, we identified 47 papers. After applying the above filtering criteria, we narrowed this down to 41 relevant papers.

The excluded literature is as follows:

\begin{itemize}

\item \textbf{An Agent-Based Model of COVID-19 Diffusion to Plan and Evaluate Intervention Policies} \cite{pescarmona2021agentbasedmodelcovid19diffusion}: An agent-based model demonstrating effective reduction of symptomatic COVID-19 cases through targeted measures and genetic algorithms. (\textit{\textbf{Note}}: This is the 2021 paper shown in Fig \ref{fig:chart_paper_num}.)

\item \textbf{PLANNER: Generating Diversified Paragraphs via Latent Language Diffusion Model} \cite{NEURIPS2023_fdba5e0a}: Text generation using autoregressive models and latent language diffusion, relevant to LLMs and NLP.

\item \textbf{Professional Basketball Player Behavior Synthesis via Planning with Diffusion} \cite{chen2023professionalbasketballplayerbehavior}: Sports analytics planning using a diffusion probabilistic model, not within our scope.

\item \textbf{Denoising Heat-inspired Diffusion with Insulators for Collision-Free Motion Planning} \cite{chang2023denoising}: Motion planning using collision-avoiding diffusion kernels and visual input for obstacle detection, not relevant to our review.

\item \textbf{First measurements of radon-220 diffusion in mice tumors, towards treatment planning in diffusing alpha-emitters radiation therapy} \cite{heger2024measurementsradon220diffusionmice}: First estimates of radon-220 diffusion lengths in mice tumors (0.25-0.6 mm), crucial for treatment planning in diffusing alpha-emitters radiation therapy.

\item \textbf{Mastering Text-to-Image Diffusion: Recaptioning, Planning, and Generating with Multimodal LLMs} \cite{yang2024masteringtexttoimagediffusionrecaptioning}: The RPG framework enhances text-to-image diffusion models using multimodal LLMs, enabling precise image generation and editing without additional training.

\end{itemize}

\definecolor{mylightpurple}{rgb}{0.9, 0.8, 1} 
\definecolor{mylightblue}{rgb}{0.8, 0.9, 1} 
\definecolor{mylightgreen}{rgb}{0.8, 1, 0.8} 

\ifthenelse{\boolean{is_access}}{

\begin{figure*}[htbp]
    \centering
    \includegraphics[width=\textwidth]{paper/figures/tree.png}
    \caption{Overview of diffusion model category for planning}
    \label{fig:chart_paper_num}
\end{figure*}

}{

\begin{figure*}[h]
    \centering
    \begin{forest}
    for tree={
        grow=east,
        reversed=true, 
        draw,
        rectangle,
        rounded corners,
        align=center,
        minimum width=5em,
        minimum height=1.5em,
        edge={thick},
        font=\normalsize,
        inner sep=2pt,
        inner xsep=5pt,
        s sep=3pt,
        child anchor=west,
        parent anchor=east,
        anchor=west,
        fill=mylightpurple!20, 
        edge path={
            \noexpand\path [draw, \forestoption{edge}] (!u.parent anchor) -- (.child anchor)\forestoption{edge label};
        },
    }
        [DM4Planning
            [{\hyperref[sec:general]{\secfond}}
                [{\hyperref[sec:found]{\secfondfd}} \cite{janner2022planning, wang2023diffusion}]
                [{\hyperref[sec:basics]{\secfondgeneral}} \cite{saha2023edmp, carvalho2024motion, 10161324, shan2024contrastive, chi2024diffusion, liu2024dipper}]
                [{\hyperref[sec:efficiency]{\secfondefficiency}} \cite{li2023efficient, brehmer2023edgi, hong2023diffused, dong2024diffuserlite, chen2024simple, wu2024diffusionreinforcement}]
                [{\hyperref[sec:pref-custom]{\secfondprefer}} \cite{dong2024aligndiff}]
                [{\hyperref[sec:replan]{\secfondreplan}} \cite{zhou2023adaptive}]
            ]
            [{\hyperref[sec:skillcond]{\secskillcond}}
                [{\hyperref[sec:skill]{\secskill}} \cite{xu2023xskill, mishra2023generative, liang2024skilldiffuser}]
                [{\hyperref[sec:cond]{\seccond}} \cite{ni2023metadiffuser, ajay2023conditional, liang2023adaptdiffuser, kondo2024cgd}]
            ]
            [{\hyperref[sec:rob]{\secrob}}
                [{\hyperref[sec:safety]{\secrobsafe}} \cite{botteghi2023trajectory, lee2023refining, xiao2023safediffuser, wang2023cold, feng2024ltldog}]
                [{\hyperref[sec:uncertainty]{\secrobuncertain}} \cite{yang2023planning, cachay2023dyffusion, NEURIPS2023_fe318a2b, fang2023dimsam}]
            ]
            [{\hyperref[sec:app]{\secapp}}
                [{\hyperref[sec:3d]{\secapptd}} \cite{huang2023diffusionbased, ze20243d, pan2024exploiting}]
                [{\hyperref[sec:inst-video]{\secappvideo}} \cite{wang2023pdppprojected, fang2023masked, shi2024actiondiffusion}]
                [{\hyperref[sec:autonomous-d]{\secappav}} \cite{yang2024diffusiones}]
                [{\hyperref[sec:mult-task]{\secappmult}} \cite{he2023diffusion}]
            ]
        ]
    \end{forest}
    \caption{Overview of literature categorization.}
    \label{fig:tree}
\end{figure*}
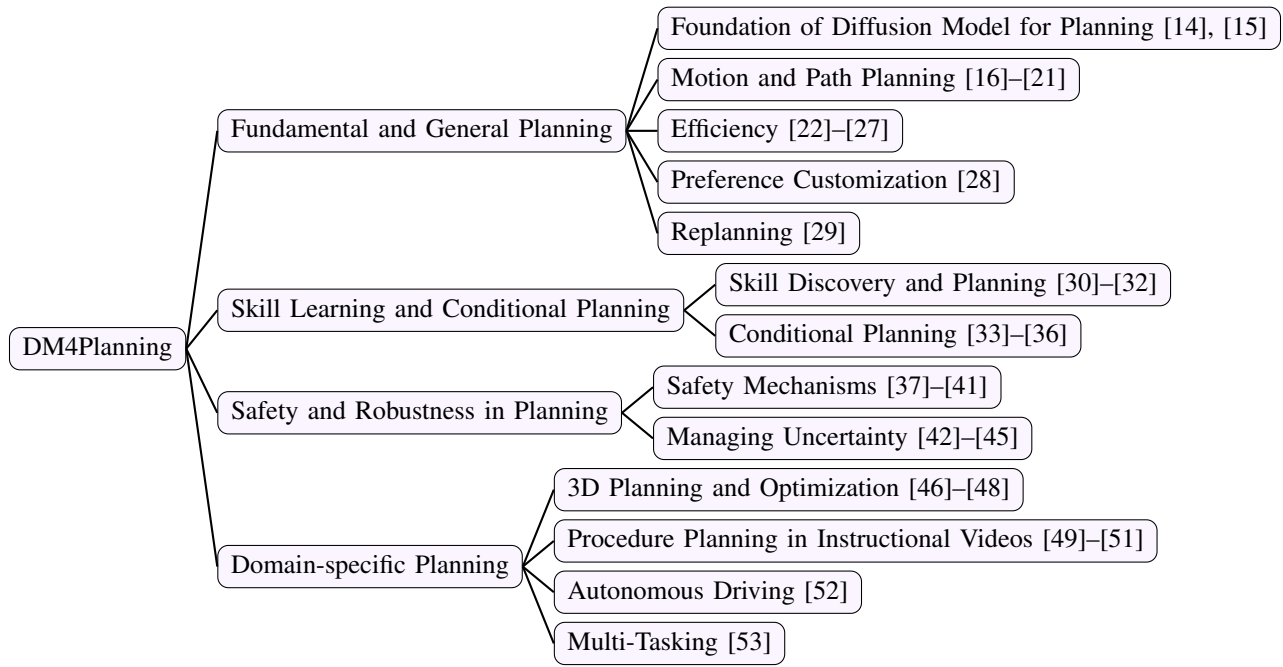
}

\section{Dataset and Benchmark} \label{sec:dataset}

The evaluation of diffusion models for planning encompasses a variety of datasets tailored to distinct tasks, ranging from motion planning and maze solving to robotic manipulation and reinforcement learning.
This section summarizes the dataset used for diffusion based planning proposals.
Each dataset presents unique challenges that highlight the models' capabilities and limitations.

\subsection{Motion and Path Planning}
\textbf{D4RL Benchmark Suite} \cite{fu2021d4rl} benchmark suite is providing standardized datasets and environments,
mainly for use in the field of offline reinforcement learning.
It includes tasks such as Maze2D, locomotion tasks like HalfCheetah, Hopper, and Walker2D, and complex multi-step manipulation tasks in kitchen environments.

\textbf{Gym-MuJoCo} \cite{1606.01540} provides simulated environments for continuous control tasks using a physics engine, including HalfCheetah, Hopper, and Walker2D.
These environments involve locomotion tasks and are used to evaluate the performance of models in dynamic movement and control settings.

\textbf{AntMaze}
AntMaze is a benchmark environment that is part of the MuJoCo suite \cite{1606.01540}.
It presents challenges in path planning and navigation within complex mazes.
This dataset is used to test models for their path-finding and navigation capabilities in constrained environments.

\textbf{Franka Kitchen} \cite{pmlr-v100-gupta20a}, \textbf{Real-world Kitchen}, and \textbf{Adroit} \cite{Rajeswaran-RSS-18} datasets involve complex manipulation tasks in both simulated and real-world environments.
Kitchen tasks include multi-step actions such as opening microwaves, moving objects, opening ovens, and grasping cloths.
Adroit tasks involve dexterous manipulation with robotic hands.
These environments are used to evaluate models like \textit{XSkill} \cite{xu2023xskill} and \textit{Generative Skill Chaining} \cite{mishra2023generative}, testing the models' abilities to handle complex sequences, precise manipulations, skill transfer, and long-horizon task planning capabilities in realistic settings.

\textbf{RLBench} \cite{9001253} provides a variety of robotic tasks such as opening and closing boxes, which are used to evaluate adaptive replanning capabilities in dynamic environments.
Models like \textit{Adaptive Online Replanning} \cite{zhou2023adaptive} use these tasks to demonstrate flexibility and real-time adjustment abilities.

\textbf{Block Stacking Tasks} involve stacking blocks to achieve specified configurations and are used in evaluating models.
These tasks assess the models' precision in manipulation and planning under physical constraints.

\subsection{3D Planning}
Datasets such as PROX \cite{DBLP:conf/iccv/HassanCTB19} and ScanNet \cite{DBLP:conf/cvpr/DaiCSHFN17} are used for 3D scene understanding, human pose generation, grasp planning, and robot arm motion planning.
These datasets test the integration of scene-conditioned generation, optimization, and goal-oriented planning in 3D environments.

\subsection{Instructional Video}
Instructional video datasets like CrossTask \cite{DBLP:conf/cvpr/ZhukovACFLS19} and COIN \cite{DBLP:conf/cvpr/TangDRZZZL019} focus on generating action sequences from start to goal visual observations.
Models like \textit{PDPP} \cite{wang2023pdppprojected} and \textit{ActionDiffusion} \cite{shi2024actiondiffusion} use these datasets to evaluate their ability to plan and execute complex procedures based on visual and language instructions.

\subsection{Autonomous Driving}
The nuPlan \cite{nuplan} dataset is a benchmark for closed-loop planning in autonomous driving.
It assesses models like \textit{Diffusion-ES} \cite{yang2024diffusiones} for optimizing non-differentiable reward functions and following complex instructions in real-time decision-making and trajectory optimization for autonomous vehicles.

\subsection{Safety-Critical Datasets}
Safety-critical datasets introduce constraints in experiments using datasets like Maze2D and Gym-MuJoCo to assess the models' ability to ensure safety and feasibility in trajectory planning. 
These datasets are used for evaluating models like \textit{SafeDiffuser} \cite{xiao2023safediffuser} and \textit{Cold Diffusion on the Replay Buffer} \cite{wang2023cold}.
The main purpose is to test the safety and robustness of the models in generating reliable and feasible plans in various environments.

\mySummaryBoxContent{
\textbf{Summary - \secdataset}\\
The datasets used for evaluating diffusion-based planners vary, including motion planning, maze solving, and robotic tasks.
Specifically, datasets like D4RL, Gym-MuJoCo, and AntMaze, are used to evaluate the models' capabilities in control, navigation, and long-horizon planning.
Complex manipulation challenges are addressed with Kitchen and Adroit datasets, while RLBench and block stacking tasks focus on adaptive replanning and precision.
The Franka Kitchen datasets test skill transfer in realistic environments.
For 3D planning, datasets like PROX and ScanNet evaluate scene understanding and optimization.
Additionally, instructional video datasets and the nuPlan dataset for autonomous driving assess the models' ability to follow procedural instructions and make real-time decisions.
Safety-critical datasets underscore the importance of robustness and reliability in trajectory planning.
}

\section{\secfond}\label{sec:general}

\begin{table*}[htbp]
\centering
\caption{Detailed key metrics for researches introduced in \secfond}
\label{tab:fond}
\begin{tabular}{c p{5cm} p{5cm} c p{4cm}}
\toprule
\textbf{Study} & \textbf{Target problem} & \textbf{Usage of diffusion model} & \textbf{Year} & \textbf{Baseline} \\
\midrule

\cite{janner2022planning}           & First paper diffusion model for planning                      & Trajectory planning                                                     & 2022 & CQL, IQL, MPPI, BCQ                                                                                                           \\
\cite{wang2023diffusion}            & First Offline Reinforcement Learning based on diffusion model & Policy representation                                                   & 2023 & BC, AWAC, Diffuser, MoRel, Onestep RL, TD3+BC, DT, CQL, IQL                                                                   \\
\cite{saha2023edmp}                 & Motion Planning for Robotic Manipulation                      & Cost functions to improve trajectory diversity and success              & 2023 & CHOMP, OMPL, G. Fabrics, STORM, M$\pi$Nets, MPNets                                                                                \\
\cite{carvalho2024motion}           & Motion Planning for Robotics                                  & Merging prior sampling and motion optimization                          & 2024 & RRTConnect, GPMP, CVAE, Stochastic-GPMP, DiffusionPrior, CVAEPosterior                                                        \\
\cite{10161324}                     & Path Planning for Mobile Sensor Networks                      & Adding random perturbations to gradient flow dynamics                   & 2024 & APF, MPC, PRM, RRT                                                                                                            \\
\cite{shan2024contrastive}          & RL for long-term planning                                     & Using contrastive learning                                              & 2023 & CQL, IQL, Decision Transformer, Trajectory Transformer, MOPO, Diffuser, Decision Diffuser                                     \\
\cite{chi2024diffusion}             & Robot Visuomotor Policy Learning                              & Control Robot visuomotor for high-frequency action tasks                & 2024 & LSTM-GMM, IBC, BET                                                                                                            \\
\cite{liu2024dipper}                & 2D Path Planning for Legged Robots                            & Image-conditioned planner                                               & 2024 & A*, N-A*, ViT-A*                                                                                                              \\
\cite{stamatopoulou2024dippest}     & Path Planning on Quadruped Robots                             & Zero-shot image and goal-conditioned planning                           & 2024 & DiPPeR, IPlanner, NoMad                                                                                                       \\
\cite{li2023efficient}              & Offline Reinforcement Learning with latent action spaces      & Continuous latent action space representation                           & 2023 & CQL, IQL, Decision Transformer, Trajectory Transformer, MoReL, Diffuser, Decision Diffuser, TAP, HDMI, D-QL                   \\
\cite{brehmer2023edgi}              & Model-Based Reinforcement Learning                            & Planning for spatial, temporal, and permutation symmetries              & 2023 & BCQ, CQL, Diffuser                                                                                                            \\
\cite{dong2024diffuserlite}         & Real-time Diffusion Planning for Reinforcement Learning       & Increasing decision-making frequency and reducing redundant information & 2024 & Diffuser, Decision Diffuser, AlignDiff, DD-small                                                                              \\
\cite{dong2024aligndiff}            & Aligning Diverse Human Preferences in RL                      & Leveraging RLHF for zero-shot behavior customization                    & 2024 & Goal Conditioned Behavior Cloning, Sequence Modeling, TD                                                                      \\
\cite{hong2023diffused}             & Multi-task Decision-making and Long-term Planning             & Planing milestones in a latent space for efficient long-term planning   & 2023 & CQL, IQL, ContRL, Decision Transformer (DT), Trajectory Transformer (TT), Decision Diffuser (DD)                              \\
\cite{chen2024simple}               & Hierarchical Planning with Diffusion Models                   & Generating sparse subgoal for larger receptive                          & 2024 & MPPI, IQL, Diffuser, IRIS, HiGoC, HDMI, BCQ, BEAR, CQL, Decision Transformer (DT), MoReL, Trajectory Transformer (TT), RvS-G  \\
\cite{wu2024diffusionreinforcement} & Hierarchical Motion Planning in Adversarial Multi-agent Games & Planing global paths considering static map constraints                 & 2024 & A-Star Heuristic, RRT-Star Heuristic, VO Method, DDPG, SAC, Diffusion Only                                                    \\
\cite{zhou2023adaptive}             & Adaptive Online Replanning                                    & Generating and refine trajectories while determining when to replan     & 2023 & BCQ, CQL, IQL, DD                                                                                                             \\

\bottomrule
\end{tabular}
\end{table*}

\subsection{\secfondfd}\label{sec:found}
This section introduces the foundational concepts that have studied the current landscape of planning with diffusion models.

\textit{Diffuser} \cite{janner2022planning} stands as a pioneering study in this field.
This model is positioned within model-based reinforcement learning, integrating trajectory optimization directly into the model learning process through diffusion models.

It introduces a diffusion probabilistic model designed for trajectory planning, which predicts all timesteps concurrently.
This approach is significant for its innovative use of iterative denoising processes, allowing for flexible conditioning and planning.
Additionally, the employment of auxiliary guides to adjust sampling strategies enhances the model's ability to satisfy high returns or constraints.
The \textit{Diffuser}'s contributions lie in its exploration of diffusion-based planning methods, demonstrating robust strategies like classifier-guided sampling and image inpainting to address long-horizon decision-making and adaptability during test time.

On the other hand,  \textit{Diffusion-QL} \cite{wang2023diffusion} focuses on offline reinforcement learning, presenting a conditional diffusion model for policy representation.
This model leverages the expressiveness of diffusion processes to capture complex, multi-modal action distributions, which are essential for offline RL tasks.
By combining behavior cloning with Q-learning guidance during the diffusion model training, \textit{Diffusion-QL} effectively generates high-value actions through an iterative denoising process guided by Q-values.
This method showcases its superiority by outperforming previous techniques in various benchmark scenarios, highlighting the potential of diffusion models to enhance policy representation and action quality in offline settings.

The evolution of these technologies underscores a common reliance on diffusion processes to facilitate flexible and efficient planning and decision-making.
Both approaches emphasize iterative denoising as a core technical idea, enabling models to handle complex prediction and optimization tasks effectively.
While \textit{Diffuser} \cite{janner2022planning} focuses on trajectory-level planning within a model-based RL framework, \textit{Diffusion-QL} \cite{wang2023diffusion} extends the utility of diffusion models to policy representation in offline RL.

\subsection{\secfondgeneral}\label{sec:basics}
Targeting motion and path planning, this section introduces diverse approaches leveraging diffusion models, comparing their contributions, proposed methodologies.

\textit{Ensemble-of-costs-guided Diffusion for Motion Planning (EDMP)} \cite{saha2023edmp} focuses on motion planning for robotic manipulation, integrating classical and deep-learning approaches.
It demonstrates generalization to diverse and out-of-distribution scenes, effectively generating multimodal trajectories.
Its concept is the utilization of a diffusion-based network to learn priors over kinematically valid trajectories, incorporating scene-specific costs through an ensemble of cost functions to enhance trajectory diversity and success rates.

In contrast, \textit{Intermittent Diffusion Based Path Planning} \cite{10161324}
caters to heterogeneous mobile sensors navigating cluttered environments.
This method introduces random perturbations to gradient flow dynamics, aiding sensors in escaping local minima.
A decentralized approach leverages local information for collision avoidance, bypassing the offline planning phase and ensuring smooth navigation via a projection strategy.

\textit{Contrastive Diffuser} \cite{shan2024contrastive} shifts focus to reinforcement learning for long-term planning.
By employing contrastive learning, it refines the base distribution of diffusion-based RL methods.
The return contrast mechanism directs generated trajectories towards high-return states, enhancing performance across multiple benchmarks.
This method exemplifies the efficacy of combining diffusion models with reinforcement learning principles.

\textit{Motion Planning Diffusion}\cite{carvalho2024motion} innovates in robot motion planning by integrating trajectory generative models with optimization-based planning.
The approach involves sampling from the posterior distribution using diffusion models, accelerating motion planning and improving solution quality.
The temporal U-Net plays a pivotal role in encoding the diffusion model over trajectories.

For legged robots, \textit{DiPPeR (Diffusion-based 2D Path Planner)} \cite{liu2024dipper} emphasizes scalability and speed.
This framework employs an image-conditioned diffusion planner and a robust training pipeline using CNNs, achieving faster trajectory generation compared to traditional and data-driven methods.
Its efficacy is validated through deployment on real-world robotic platforms like Boston Dynamics’ Spot and Unitree’s Go1.

\textit{Diffusion Policy}\cite{chi2024diffusion} advances robot visuomotor control, utilizing a conditional denoising diffusion process.
Key technical contributions include receding horizon control, visual conditioning, and a time-series diffusion transformer.
These innovations enable robust execution across multiple manipulation tasks, outperforming several baseline methods.

Building on previous research into A* and RRT* heuristics \cite{ye2023learning}, the study by Wu \textit{et al.} \cite{wu2024diffusionreinforcement} presents a hierarchical framework that integrates diffusion models with reinforcement learning for evasive planning.
This dual-layer architecture features a high-level diffusion model for global path planning and low-level reinforcement learning algorithms for local evasive maneuvers.
Task-oriented costmaps enhance the framework by improving explainability and predictability through the consideration of detection risks and dynamic adjustments.

Finally, \textit{DiPPeST (Diffusion-based Path Planner for Synthesizing Trajectories)} \cite{stamatopoulou2024dippest} extends the capabilities of diffusion-based planning to quadrupedal robots.
The proposal uses RGB input for global path generation and real-time refinement, achieving efficient navigation and obstacle avoidance without additional training.
The integration with a visual  framework underscores its practical applicability in real-world scenarios.

\subsection{\secfondefficiency}\label{sec:efficiency}
This section explores improving computational efficiency and decision-making effectiveness in diffusion model-based planning.

\textit{Latent Diffuser }\cite{li2023efficient} addresses offline reinforcement learning by proposing a framework for continuous latent action space representation.
This approach leverages latent, score-based diffusion models, showcasing the theoretical equivalence between planning in the latent action space and energy-guided sampling with a pre-trained diffusion model.
The introduction of a sequence-level exact sampling method ensures competitive performance in both low-dimensional locomotion control and higher-dimensional tasks, thereby advancing the state of efficient planning.

\textit{Equivariant Diffuser for Generating Interactions (EDGI)} \cite{brehmer2023edgi} further pushes the boundaries in model-based reinforcement learning.
By incorporating spatial, temporal, and permutation symmetries through an SE(3) × Z × Sn-equivariant diffusion model, EDGI enhances sample efficiency and generalization.
This method also introduces flexible soft symmetry breaking during test time for task-specific adaptations, thereby improving planning efficiency and the ability to generalize across different symmetry groups.

In the realm of multi-task decision-making and long-term planning, \textit{Diffused Task-Agnostic Milestone Planner (DTAMP)} \cite{hong2023diffused} utilizes diffusion-based generative sequence models, DTAMP effectively handles long-horizon tasks and vision-based control.
It achieves state-of-the-art performance on D4RL and CALVIN benchmarks by planning milestones in a latent space and using goal-conditioned imitation learning.
The implementation of classifier-free diffusion guidance ensures efficient path planning, highlighting its robustness in multi-task environments.

\textit{DiffuserLite} \cite{dong2024diffuserlite} introduces a real-time diffusion planning approach using its plan refinement process (PRP), which generates coarse-to-fine-grained trajectories to minimize redundant information.
Its ability to function as a flexible plugin with other diffusion planning algorithms marks a leap toward real-time diffusion planning.

Finally, \textit{Hierarchical Diffuser} \cite{chen2024simple} introduces a framework that combines hierarchical and diffusion-based planning to improve efficiency in long-horizon tasks.
By employing a high-level diffuser for sparse subgoal generation and a low-level diffuser for detailed subgoal achievement, this approach enhances computational efficiency and generalization capabilities.
The introduction of \textit{Sparse Diffuser} and \textit{Sparse Diffuser with Dense Actions (SD-DA)} variants further improves return prediction and task performance.

\subsection{\secfondprefer}\label{sec:pref-custom}
For human preference alignment in reinforcement learning, 
\textit{AlignDiff} \cite{dong2024aligndiff} introduces a framework combining reinforcement learning from human feedback (RLHF) with diffusion models for zero-shot behavior customization, addressing challenges in aligning agent behaviors with abstract and mutable human preferences.
Key contributions are using RLHF to quantify multi-perspective human feedback datasets and developing metrics to evaluate agents' preference matching, switching, and covering capabilities, demonstrating superior performance.
The proposal involves a transformer-based attribute strength model for quantifying behavior strengths and an attribute-conditioned diffusion model acting as a planner, guided by the attribute strength model during inference.
Baseline methods include goal-conditioned behavior cloning and so on.
It is evaluated using diverse locomotion tasks in simulation environments like Hopper, Walker, and Humanoid from benchmarks such as MuJoCo and DMControl.

\subsection{\secfondreplan}\label{sec:replan}
Under unforeseen and changing conditions, replanning is crucial for enhancing adaptability and robustness in dynamic environments. 
\textit{Replanning with Diffusion Models (RDM)} \cite{zhou2023adaptive} introduces an approach to determine optimal replanning times, enhancing planning efficiency and effectiveness in stochastic and long-horizon tasks.
A key contribution is the introduction of a principled method that assesses the likelihood of the current plan's executability.
This method decides when to replan to ensure trajectories remain consistent with the original goal state.
The approach includes multiple replanning strategies: replanning from scratch, replanning from previous context, and replanning with future context.
This highlights the robustness and broad applicability in complex scenarios, representing an advancement in adaptive replanning for diffusion models.

\mySummaryBoxContent{
\textbf{Summary - \secfond}\\
Diffusion models enhance various aspects of reinforcement learning, including state and policy representation and trajectory optimization.
Foundational approaches integrate trajectory optimization directly into model-based RL and employ conditional diffusion models for policy representation in offline RL, demonstrating superior performance in generating high-value actions through iterative denoising processes.
For motion and path planning, diffusion models improve precision and efficiency in complex tasks, such as robotic manipulation and mobile sensor navigation.
Efficiency enhancements are seen in methods utilizing continuous latent action spaces and equivariant diffusion models for better generalization.
Additionally, combining reinforcement learning from human feedback with diffusion models enables zero-shot behavior customization, and adaptive replanning strategies ensure robust planning in dynamic environments.
Overall, diffusion models advance RL by improving flexibility, efficiency, and performance in diverse applications.
}

\section{\secskillcond}\label{sec:skillcond}
In this section, we delves into the state of the art in skill discovery, task planning, and trajectory generation through the lens of diffusion models and conditional generative approaches.
These methodologies offer solutions to complex problems, ranging from cross-embodiment skill transfer to adaptive planning in dynamic environments.

The papers introduced in this section are summarized in Table \ref{tab:skill}, highlighting their target problems and the application of diffusion models within these fields.
\begin{table*}[htbp]
\centering
\caption{Detailed key metrics for researches introduced in \secskillcond}
\label{tab:skill}
\begin{tabular}{c p{5cm} p{7cm} c p{2cm}}
\toprule
\textbf{Study} & \textbf{Target problem} & \textbf{Usage of diffusion model} & \textbf{Year} & \textbf{Dataset} \\
\midrule

\cite{xu2023xskill}           & Cross-Embodiment Skill Discovery           & Translates observed human demonstrations into robot actions        & 2023 &  Franka Kitchen, Realworld Kitchen \\
\cite{mishra2023generative}   & Long-Horizon Task and Motion Planning      & Generate plans adhere to specified constraints                     & 2023 &  PyBullet, Franka Panda arm        \\
\cite{liang2024skilldiffuser} & Hierarchical Planning and Diffusion Models & Generate skill-oriented state conditioned on learned skills        & 2024 &  LOReL Sawyer, Meta-World          \\
\cite{ni2023metadiffuser}     & Offline Meta RL                            & Generating task-specific trajectories using contextual information & 2023 &  MuJoCo                            \\
\cite{ajay2023conditional}    & Offline Decision Making                    & Generate optimal trajectories without the dynamic programming      & 2023 &  D4RL                              \\
\cite{liang2023adaptdiffuser} & Offline RL                                 & Generating and utilizing synthetic data guided by reward gradients & 2023 &  D4RL                              \\
\cite{kondo2024cgd}           & UAV Trajectory Planning                    & Plan with dynamically adjustment to new constraints                & 2024 &  Simulated UAV flight scenarios    \\

\bottomrule
\end{tabular}
\end{table*}

\subsection{\secskill}\label{sec:skill}
Skill Discovery and Planning section focus on extracting and leveraging skills from both human and robot demonstrations, enabling robots to understand and replicate complex tasks through techniques like imitation learning, hierarchical planning, and generative models.
This allows for efficient task execution based on skill abstraction and diffusion models.

\textit{XSkill} \cite{xu2023xskill}, positioned as a cross-embodiment skill discovery framework, leverages imitation learning to bridge the gap between human and robot demonstrations.
By developing a self-supervised algorithm to extract skill prototypes from unstructured videos, this work introduces a skill-conditioned diffusion policy that translates human demonstrations into robot actions.
Furthermore, it incorporates a skill alignment transformer to align and compose learned skills for executing unseen tasks based on human prompts.
This study is pivotal in addressing the challenge of transferring skills across different embodiments, ensuring robots can perform tasks demonstrated by humans despite inherent physical differences.

\textit{Generative Skill Chaining (GSC)} \cite{mishra2023generative} tackles the complexities of long-horizon task and motion planning by integrating generative modeling with skill chaining.
Their main proposal lies in its probabilistic framework that captures the joint distribution of skill preconditions, parameters, and effects.
This approach allows for efficient generation of skill sequences through forward and backward diffusion processes, ensuring both initial state and final goal conditions are met.
Additionally, \textit{GSC} incorporates constraint-based planning to enhance the feasibility of generated plans, making it a scalable solution for complex, multi-step tasks that require dynamic adaptation.

\textit{SkillDiffuser} \cite{liang2024skilldiffuser} offers an interpretable hierarchical planning framework, combining skill abstractions with diffusion-based task execution.
It proposes an end-to-end approach where discrete, interpretable skills are learned and used to condition a diffusion model for trajectory planning.
By utilizing a classifier-free guidance model, it generates state trajectories that align with high-level skill abstractions derived from visual and language inputs.
This framework excels in providing interpretable visualizations of skills, ensuring robust and adaptable performance across multiple tasks, including those with ambiguous language instructions.

The development paths of these technologies illustrate distinct but complementary approaches; 
\textit{XSkill} \cite{xu2023xskill} emphasizes cross-embodiment skill transfer through self-supervised learning and skill-conditioned policies.
\textit{GSC} \cite{mishra2023generative} focuses on long-horizon planning with a probabilistic approach to skill chaining, leveraging forward and backward diffusion processes.
\textit{SkillDiffuser} \cite{liang2024skilldiffuser} integrates hierarchical planning with interpretable skill abstractions, utilizing diffusion models to generate coherent state trajectories.

\subsection{\seccond}\label{sec:cond}
This section emphasizes generating task-specific trajectories and policies by conditioning on various contextual inputs, such as task requirements and environmental constraints.
This approach uses diffusion models and other generative techniques to create feasible and optimal plans from offline data, thereby enhancing decision-making and adaptability in diverse and dynamic scenarios.

\textit{MetaDiffuser} \cite{ni2023metadiffuser} stands out in the realm of offline meta-reinforcement learning by introducing a context-conditioned diffusion model for task-oriented trajectory generation.
Its dual-guided module promotes both dynamic consistency and high returns, demonstrating superior generalization across diverse unseen tasks without environment interaction during the learning process.
The proposal to integrate a context encoder with a conditional diffusion model allows for the generation of trajectories tailored to specific tasks, enhancing dynamic feasibility and high-return outcomes.

In contrast,  \textit{Decision Diffuser} \cite{ajay2023conditional} positions conditional generative modeling as a viable alternative to traditional reinforcement learning methods. This approach leverages a return-conditional diffusion model to simplify policy generation by avoiding value function estimation, thus bypassing the complexities of dynamic programming.
The flexibility of this model in handling various constraints and skill compositions during inference is particularly notable, highlighting its broader applicability in decision-making tasks.

\textit{AdaptDiffuser} \cite{liang2023adaptdiffuser} further extends the diffusion model-based planning methods by incorporating an evolutionary approach.
It generates and utilizes synthetic data guided by reward gradients to fine-tune the diffusion model, enhancing adaptability and performance on both seen and unseen tasks.
This framework's innovative use of gradient-based guidance during the generation process ensures that the data produced aligns closely with task-specific requirements, providing substantial performance improvements over prior methodologies.

Lastly, \textit{Constraint-Guided Diffusion Policies (CDG)} \cite{kondo2024cgd} addresses UAV trajectory planning through an imitation learning-based approach.
By dividing the trajectory planning problem into sub-problems of collision avoidance and dynamic feasibility, CGD employs block-coordinate descent to iteratively optimize trajectory parameters.
This method demonstrates remarkable adaptability to new constraints, ensuring dynamically feasible and collision-free UAV trajectories in real-time scenarios.

Comparatively, \textit{MetaDiffuser} \cite{ni2023metadiffuser} and \textit{AdaptDiffuser} \cite{liang2023adaptdiffuser} emphasize enhancing generalization and adaptability through innovative uses of synthetic data and dual-guided modules.
\textit{Decision Diffuser} \cite{ajay2023conditional} approach simplifies policy generation and showcases flexibility in decision-making contexts.
Meanwhile, \textit{CDG} \cite{kondo2024cgd} focuses on real-time adaptability and constraint handling in UAV planning.

\mySummaryBoxContent{
\textbf{Summary - \secskillcond}\\
In this section, we explored advancements in skill learning and conditional planning using diffusion models.
These innovations focus on various aspects such as cross-embodiment skill discovery, long-horizon task and motion planning, hierarchical planning, offline meta-reinforcement learning, and UAV trajectory planning.
The methodologies presented leverage the power of diffusion models to enhance the flexibility, scalability, and generalization capabilities of robotic systems.
By integrating skill learning with conditional planning, these approaches address complex planning problems, ensuring robust and adaptive performance across diverse environments and tasks.
}

\section{\secrob}\label{sec:rob}
This section explores the safety and robustness of diffusion model planning, building on existing research in offline reinforcement learning robustness, such as the work by Panaganti \textit{et al.} \cite{panaganti2022robust}.
First, we discuss methods the safety mechanisms for diffusion-based planning, ensuring the generation of safe plans.
Next, we address approaches that manage uncertainty and partial observations, improving the adaptability and accuracy of diffusion models in dynamic environments.
The papers introduced in this section are shown in Table \ref{tab:rob}, which explains their target problems, usage of diffusion models.

\begin{table*}[htbp]
\centering
\caption{Detailed key metrics for researches introduced in \secrob}
\label{tab:rob}
\begin{tabular}{c p{5cm} p{6.5cm} c p{2.5cm}}
\toprule
\textbf{Study} & \textbf{Target problem} & \textbf{Usage of diffusion model} & \textbf{Year} & \textbf{Dataset} \\
\midrule

\cite{botteghi2023trajectory} & Trajectory Generation                                           & DDPMs with control barrier functions                              & 2023 & OpenAI Gym                                                \\
\cite{lee2023refining}        & Reliable behavior synthesis and planning                        & Generating "restoration gap" to evaluate the quality of plans     & 2023 & D4RL, Kuka iiwa robotic arm                               \\
\cite{xiao2023safediffuser}   & Safe trajectory planning in safety-critical applications        & Applying a quadratic program in each diffusion step for safety    & 2023 & MuJoCo                                                    \\
\cite{wang2023cold}           & Learning from Demonstrations (LfD) and trajectory planning      & Guidance  through the replay buffer of previously visited states  & 2023 & Maze, Fetch robot environments                            \\
\cite{feng2024ltldog}         & Safe trajectory planning with temporal and symbolic constraints & Ensuring adherence to finite linear temporal logic constraints    & 2024 & D4RL, PushT, Real Robot Environments                      \\
\cite{yang2023planning}       & Embodied task planning under uncertainty                        & Jointly model state trajectory and goal estimation                & 2023 & Kuka Robot, CompILE, ALFRED                               \\
\cite{cachay2023dyffusion}    & Probabilistic spatiotemporal forecasting                        & Reducing computational complexity incorporating temporal dynamics & 2023 & Sea Surface Temperatures, Navier-Stokes flow, Spring Mesh \\
\cite{NEURIPS2023_fe318a2b}   & Uncertainty-Aware Planning                                      & Conformal prediction-based uncertainty estimation                 & 2023 & D4RL                                                      \\
\cite{fang2023dimsam}         & Task and Motion Planning                                        & Learning constraint-satisfying samplers for TAMP                  & 2023 & IsaacGym, Point clouds from partial observations          \\

\bottomrule
\end{tabular}
\end{table*}

\subsection{\secrobsafe}\label{sec:safety}

In the domain of safety mechanisms, the integration of DDPMs with control barrier functions (CBFs) has been a development \cite{botteghi2023trajectory}.
The framework developed for safety-critical optimal control employs three distinct DDPMs for dynamically consistent trajectory generation, value estimation, and safety classification.
This method introduces a guided sampling procedure for generating optimal and safe trajectories, leveraging conditional sampling schemes that combine value function and safety classifier guides.
The primary contribution lies in combining performance and safety through modular components that can be retrained for new environments, promising scalability and efficiency enhancements.

Another contribution focuses on refining the reliability of plans generated by diffusion models \cite{lee2023refining}.
The introduction of the "restoration gap" metric evaluates the quality of generated plans, coupled with a gap predictor for refinement.
This approach ensures more feasible and reliable plans by guiding the reduction of the restoration gap and mitigating adversarial guidance through attribution map regularization.
This method demonstrates its effectiveness across different benchmarks, emphasizing its potential for real-world, safety-critical applications.

By incorporating finite-time diffusion invariance and embedding control barrier functions into the diffusion process, \textit{SafeDiffuser} \cite{xiao2023safediffuser} ensures safety constraints are consistently met.
The development of different variants like Robust-Safe, Relaxed-Safe, and Time-Varying-Safe Diffusers highlights its adaptability to various safety-critical tasks, demonstrating robustness and effectiveness in diverse environments such as maze planning and robot locomotion.

Further advancing planning reliability,  \textit{Cold Diffusion on the Replay Buffer (CDRB)} \cite{wang2023cold} utilizes cold diffusion to optimize the planning process via an agent's replay buffer of previously visited states.
This method improves obstacle avoidance during planning tasks, utilizing k-means clustering to manage buffer size efficiently, thereby enhancing computational efficiency without compromising performance.

In terms of satisfying temporal and symbolic constraints, \textit{LTLDoG} \cite{feng2024ltldog} modifies the inference steps of the reverse process based on finite linear temporal logic (LTLf) instructions.
By introducing a satisfaction value function on LTLf and employing differentiable evaluation and regressor-guidance neural networks, \textit{LTLDoG} ensures the adherence to static and temporal safety constraints during trajectory generation.
This framework demonstrates its utility in robot navigation and manipulation tasks, underscoring its potential for complex and dynamic environments.

In summary, while DDPMs and their integration with CBFs and guided sampling address dynamic consistency and safety \cite{botteghi2023trajectory}, restoration gap metrics \cite{lee2023refining} and \textit{Cold Diffusion} \cite{wang2023cold} enhance plan feasibility and reliability.
\textit{SafeDiffuser} \cite{xiao2023safediffuser} ensures safety through control barrier functions, and \textit{LTLDoG} \cite{feng2024ltldog} focuses on temporal and symbolic constraint satisfaction.
Together, these contributions provide a robust foundation for developing safe, reliable, and efficient planning mechanisms across various applications.

\subsection{\secrobuncertain}\label{sec:uncertainty}
Understanding and managing uncertainty and partial observations are crucial in advancing plans' robustness and adaptability.

\textit{Planning as In-Painting} \cite{yang2023planning} introduces a language-conditioned diffusion model for task planning in partially observable environments.
This framework utilizes the diffusion model to jointly model state trajectory and goal estimation, facilitating on-the-fly planning by dynamically updating plans based on newly observed information.
Its effectiveness has been demonstrated in various embodied tasks, such as vision-language navigation and object manipulation, underscoring its robustness in handling environmental uncertainties.
Future research aims to improve real-world applicability through better handling of uncertainties, integration of complex language instructions, and expansion to 3D spaces.

Another advancement is \textit{DYffusion} \cite{cachay2023dyffusion}, a dynamics-informed diffusion model designed for spatiotemporal forecasting.
This model incorporates temporal dynamics directly into the diffusion process, utilizing a time-conditioned interpolator and forecaster network to enhance forecasting efficiency and accuracy.
Future directions include exploring advanced stochastic methods, developing faster inference strategies, and integrating new neural architectures for improved performance.

\textit{PlanCP} \cite{NEURIPS2023_fe318a2b} introduces conformal prediction for uncertainty-aware planning with diffusion dynamics models.
This work enhances the robustness and accuracy of trajectory predictions by incorporating uncertainty estimates, demonstrated across various planning and offline reinforcement learning tasks.
This method combines conformal prediction with diffusion models to reduce uncertainty while maintaining performance.

\textit{DiMSam} \cite{fang2023dimsam} utilizes diffusion models as samplers for task and motion planning (TAMP) under partial observability.
This approach leverages deep generative diffusion models to learn constraint-satisfying samplers for TAMP, defining constraints on a latent embedding of object states to handle unseen objects.
By integrating these samplers within a classical TAMP solver, it achieves long-horizon constraint-based reasoning, demonstrated on articulated object manipulation tasks.
Future work will explore advanced stochastic methods for the forward process, use more advanced ODE solvers, and apply the model to real-world problems where input and output spaces differ.

In summary, the advancements in diffusion models for planning under uncertainty reflect a diverse array of approaches;
The \textit{Planning as In-Painting} \cite{yang2023planning} focuses on dynamic task planning,
\textit{DYffusion} \cite{cachay2023dyffusion} enhances spatiotemporal forecasting through temporal dynamics,
\textit{PlanCP} \cite{NEURIPS2023_fe318a2b} integrates conformal prediction for uncertainty-aware planning, 
and \textit{DiMSam} \cite{fang2023dimsam} uses diffusion models as samplers for TAMP.
These studies improve the robustness, accuracy, and applicability of planning models in uncertain and complex environments from various perspectives.

\mySummaryBoxContent{
\textbf{Summary - \secrob}\\
The section on robustness in planning highlights the advancements in integrating safety mechanisms and managing uncertainty in diffusion-based planning models.
The discussed frameworks emphasize enhancing safety-critical applications through control barrier functions and denoising diffusion probabilistic models, ensuring optimal and safe trajectory generation.
Techniques like the \textit{CDRB} \cite{wang2023cold} and the use of restoration gap metrics refine the quality and feasibility of generated plans.
Additionally, approaches for managing uncertainty, such as the \textit{Planning as In-Painting} \cite{yang2023planning} framework and the integration of conformal prediction with diffusion dynamics models, demonstrate improvements in planning reliability under uncertain conditions.
Overall, these studies enahnce the diffusion models to deliver robust, reliable, and safe planning solutions across diverse and complex environments.
}

\section{\secapp}\label{sec:app}
The application of diffusion models is widespread in fields such as robotics, autonomous driving, and instructional content analysis.
As summarized in Table \ref{tab:app}, this section explores domain-specific planning and provides detailed insights into how diffusion models are employed to solve specific planning problems.

\begin{table*}[htbp]
\centering
\caption{Detailed key metrics for researches introduced in \secapp}
\label{tab:app}
\begin{tabular}{c p{5cm} p{5.5cm} c p{4cm}}
\toprule
\textbf{Study} & \textbf{Application} & \textbf{Usage of diffusion model} & \textbf{Year} & \textbf{Dataset} \\
\midrule

\cite{huang2023diffusionbased} & Path planning, motion planning, and optimization in 3D scenes      & Scene-conditioned generation, optimization, and planning        & 2023 & PROX, MultiDex, ScanNet, MoveIt           \\
\cite{ze20243d}                & Visual imitation learning, robot manipulation, visuomotor control  & 3D visual representations                                       & 2024 & Simulation tasks, Real robot tasks        \\
\cite{pan2024exploiting}       & View planning, object reconstruction                               & One-shot view planning from a single RGB image                  & 2024 & HomebrewedDB, UR5 robot arm in real-world \\
\cite{wang2023pdppprojected}   & Procedure planning in instructional videos                         & U-Net based distribution fitting task                           & 2023 & CrossTask, NIV, COIN                      \\
\cite{fang2023masked}          & Procedure planning in instructional videos                         & Reducing the decision space with masked diffusion               & 2023 & DDN, Ext-GAIL, PlaTe, PDPP                \\
\cite{shi2024actiondiffusion}  & Procedure planning in instructional videos                         & Generating plan incorporating action embeddings to noise mask   & 2024 & CrossTask, Coin, NIV                      \\
\cite{yang2024diffusiones}     & Autonomous driving, trajectory optimization, zero-shot instruction & Integrating gradient-free optimization and trajectory denoising & 2024 & nuPlan                                    \\
\cite{he2023diffusion}         & Multi-task RL                                                      & Generating plan and data synthesis for multi-task offline RL    & 2023 & Meta-World MT50-rand, Maze2D              \\

\bottomrule
\end{tabular}
\end{table*}

\subsection{\secapptd} \label{sec:3d}
Planning in three-dimensional environments is vital for advancements in robotics, autonomous systems, and complex scene understanding, offering solutions that bridge the gap between theoretical models and practical applications.

\textit{SceneDiffuser} \cite{huang2023diffusionbased} is a model designed for integrating scene-conditioned generation, physics-based optimization, and goal-oriented planning in 3D scenes.
Technically, it addresses the posterior collapse problem in scene-conditioned generative models through a diffusion-based denoising process.
The model uses a conditional generative approach for 3D scene understanding and proposes a unified framework that combines generation, optimization, and planning.
It employs an iterative guided-sampling framework for scene-aware generation and optimization, alongside a global trajectory planner that integrates both physics and goal awareness.
This approach facilitates path planning, motion planning, and optimization tasks in various 3D scenes.

In contrast, \textit{3D Diffusion Policy (DP3)} \cite{ze20243d} focuses on visual imitation learning and robot manipulation.
DP3 integrates 3D visual representations with diffusion policies, enabling effective visuomotor policy learning for a wide range of robotic tasks.
By employing a compact 3D visual representation extracted from sparse point clouds and an efficient MLP encoder, DP3 can handle 72 simulation tasks with minimal demonstrations, achieving high success rates in real robot tasks.
This approach highlights its generalization capabilities across various dimensions, including space, viewpoint, and instance variability.

A distinct but related approach involves using 3D diffusion models as priors for RGB-based one-shot view planning to optimize object reconstruction tasks \cite{pan2024exploiting}.
This method leverages geometric priors from 3D diffusion models to generate a 3D mesh from a single RGB image, converting the problem into a customized set covering optimization.
By applying multi-view and distance constraints, this technique effectively balances object reconstruction quality with movement cost in both simulation and real-world experiments.
The integration of NeRF-based object reconstruction further underscores the method's practical applicability.

Comparatively, \textit{SceneDiffuser} \cite{huang2023diffusionbased} and \textit{3D Diffusion Policy (DP3)} \cite{ze20243d} both rely on diffusion-based techniques, yet they diverge in their core focus;
\textit{SceneDiffuser} emphasizes scene-conditioned generative modeling and planning,
whereas \textit{DP3} excels in visuomotor policy learning with sparse demonstrations.
The RGB-based one-shot view planning approach \cite{pan2024exploiting}, while distinct in its application, shares the underlying principle of leveraging 3D diffusion models to enhance planning efficiency and effectiveness.

\subsection{\secappvideo}\label{sec:inst-video}
This section explains the advancements in leveraging diffusion models to enhance the generation of action plans from instructional content.

\textit{PDPP} \cite{wang2023pdppprojected} model introduces a novel approach to procedural planning by conceptualizing it as a conditional distribution-fitting challenge.
Utilizing a U-Net-based diffusion model, \textit{PDPP} efficiently generates action sequences that transition from initial to goal visual states.
This method's key innovations include the elimination of the need for intermediate supervision, relying solely on task labels for training.
It also introduces a projected diffusion model that enhances the efficiency of learning and sampling from the action sequence distribution.

In contrast, the \textit{Masked Diffusion with Task-awareness model} \cite{fang2023masked} advances procedural planning by integrating task-oriented attention mechanisms to manage decision spaces more effectively.
This model enhances its functionality by leveraging joint visual-text embeddings from pre-trained vision-language models, which improve task classification accuracy.
The fusion of task-aware diffusion with text-enhanced visual representation learning forms a crucial component of this model's methodology.

Meanwhile, \textit{ActionDiffusion} \cite{shi2024actiondiffusion} adopts a unique strategy by incorporating temporal dependencies between actions within its diffusion model framework to heighten the precision of action plan generation.
Its main advancements include the development of an action-aware noise mask designed to better capture these temporal dependencies and an attention mechanism within the U-Net denoising network that bolsters the learning of action correlations.
Its focus on temporal aspects distinctly enhances its ability to generate more accurate action plans.

Comparatively, \textit{PDPP} \cite{wang2023pdppprojected} distinguishes itself with its innovative projected diffusion method and its strategy of removing intermediate supervision, which simplifies the training process.
\textit{Masked Diffusion model} \cite{fang2023masked} stands out due to its effective task-aware attention mechanism, which sharpens the decision-making process and improves integration with visual-text embeddings.
Lastly, \textit{ActionDiffusion} \cite{shi2024actiondiffusion} excels through its emphasis on temporal dependencies, employing an action-aware noise mask that enhances the generation of coherent and contextually appropriate action plans.

\subsection{\secappav}\label{sec:autonomous-d}

\textit{Diffusion-ES} \cite{yang2024diffusiones} introduces Diffusion-ES, a method that combines gradient-free evolutionary search with diffusion models.
This approach optimizes trajectories for autonomous driving and instruction following on non-differentiable, black-box objectives, maintaining solution quality within the data manifold.
This work achieves state-of-the-art performance in the nuPlan autonomous driving benchmark, outperforming traditional planners and reactive policies.
It can follow complex, zero-shot natural language instructions to generate driving behaviors not present in the training data, showcasing high adaptability and generalization.
This study is evaluated against conventional sampling-based planners, deterministic and diffusion-based reactive policies, and reward-gradient-guided models.

\subsection{\secappmult}\label{sec:mult-task}
\textit{Multi-Task Diffusion Model (MTDIFF)} \cite{he2023diffusion} introduce a diffusion-based generative model designed for handling extensive multi-task offline data.
The core contribution of this research is modeling multi-task trajectory data, showcasing its ability to plan and synthesize data across various tasks; both enhance the performance and generalization of reinforcement learning systems.
The proposal underscores the potential of diffusion models in multi-task settings, advocating for further exploration to enhance generalization and adaptability.
It suggests architectural improvements to broaden task applicability and reduce computational demands, highlighting a path forward for more efficient models.
In evaluation, the authors conducted comparisons against several established models in reinforcement learning, including decision transformers and traditional multi-task reinforcement learning methods using the Meta-World and Maze2D benchmarks.

\mySummaryBoxContent{
\textbf{Summary - \secapp}\\
This section explores the diverse applications of diffusion models in planning and optimization across different domains.
It highlights advancements in 3D planning and optimization, demonstrating the integration of scene-conditioned generation with physics-based planning, as exemplified by models.
The section also examines diffusion models in visual imitation learning and robot manipulation, showcasing their ability to handle complex tasks with minimal demonstrations in scenarios like the 3D Diffusion Policy.
In view planning, diffusion models are utilized for efficient one-shot object reconstruction from RGB images.
The discussion extends to procedural planning in instructional videos, illustrating how diffusion models can generate action sequences and enhance task awareness.
Additionally, it covers the application of diffusion models in autonomous driving, focusing on trajectory optimization and instruction following.
Lastly, the potential in multi-task reinforcement learning is underscored, demonstrating their effectiveness in generative planning and data synthesis across diverse tasks.
}

\section{Research Challenges} \label{sec:challenges}
After reviewing the relevant literature in the above sections, this section discusses the future challenges.

\subsection{Combination with Other Generative Models}
Combining diffusion models with other generative models, such as Variational Autoencoders (VAEs) \cite{Kingma_2019} and Generative Adversarial Networks (GANs) \cite{goodfellow2014generative}, offers the potential for enhancing the performance, robustness, and efficiency of diffusion models used in planning.
Integrating diffusion models with VAEs can mitigate overfitting by leveraging the continuous data approximations that VAEs provide.
This allows diffusion models to generate more accurate and diverse samples, thanks to the VAEs' ability to learn latent representations that capture the underlying structure of the data \cite{xiao2023upgrading}.
Additionally, VAEs contribute to improved training stability by providing a consistent and structured latent space, which enhances the robustness of diffusion models during the training process.
Similarly, integrating diffusion models with GANs offers several benefits.
One advantage is the acceleration of inference processes, for instance, GANs have be employed to refine and enhance the outputs of diffusion models, achieving high image quality in a shorter amount of time \cite{kang2024distilling}.
Moreover, GANs help maintain fine details and sharpness in the generated images, complementing the strengths of diffusion models in capturing complex data distributions.
However, integrating these models presents challenges, particularly in ensuring model compatibility and optimization stability.
The different training objectives and model architectures of VAEs, GANs, and diffusion models necessitate careful consideration of how these models are combined.

\subsection{Scalability and Real-Time Applications} 

To achieve an efficient architecture, a critical need exists to explore advanced pruning and quantization techniques aimed at reducing the model size and computational demands without sacrificing performance \cite{DBLP:journals/corr/HanMD15}.
Furthermore, developing effective knowledge distillation methods is essential, as these methods enable the transfer of insights from larger models to more compact versions, maintaining high accuracy in resource-constrained settings.
Another challenge involves optimizing compression algorithms that dynamically adjust model complexity based on available computational resources, incorporating innovative encoding techniques to enhance storage and computational efficiency.
Integrating these optimized models with cutting-edge hardware technologies, such as GPUs and TPUs, is crucial to further boost performance and efficiency.

For real-time applications, incorporating diffusion models with edge computing would introduce unique advantages for enhancing real-time processing capabilities \cite{7488250}.
Reducing latency is a primary concern, requiring optimization of data processing pipelines and minimization of data transmission overhead between edge devices and central servers.
Ensuring the privacy and security of data processed at the edge is imperative, especially to comply with stringent data protection regulations, through advanced encryption and secure computation strategies.
Enhancing the systems' capability for real-time decision-making through algorithms optimized for low-latency execution will enable edge devices to respond swiftly and accurately to dynamic environmental inputs.

\subsection{Generalization and Robustness}

In enhancing the discussion on the challenges of generalization and robustness within advanced learning models, particularly meta-learning, key aspects from recent research highlight the complexities involved.
The Model-Agnostic Meta-Learning (MAML) approach \cite{DBLP:conf/icml/FinnAL17}, is pivotal for its ability to generalize across various tasks and models without task-specific modifications.
This highlights a core challenge: developing algorithms that not only perform reliably across diverse environments but also manage rapid adaptation without overfitting, which is a common risk with few-shot learning models.

Robustness is particularly challenged by the sensitivity of these models to initial conditions and the breadth of tasks during training.
The strategic fine-tuning from a specifically trained initial model as in MAML \cite{DBLP:conf/icml/FinnAL17}, and the insights from the work by Rahaman \textit{et.al.} \cite{DBLP:conf/icml/RahamanBADLHBC19}, underscore how slight variations in training conditions can lead to deviations in model performance when applied to new tasks.
These studies expose the delicate balance between rapid adaptation and the prevention of overfitting, emphasizing the difficulty in achieving robust generalization.

To address these challenges, cross-domain validation is proposed, ensuring models can adapt beyond their training data and not simply memorize solutions.
Furthermore, integrating model-agnostic frameworks with task-specific adjustments could foster the development of models that are both robust and adaptable.
This might involve hybrid approaches that combine meta-learning strategies with mechanisms that adjust to task-specific nuances, enhancing both the robustness and adaptive capabilities of learning models.

\subsection{Human-Robot Interaction} 

For the Human-Robot Interaction using diffusion models for planning, it would be beneficial to explore how these models might enhance the interpretability and controllability of robotic actions, crucial for effective collaboration.
Recent advancements, particularly the Diffusion Co-policy \cite{DBLP:journals/ral/NgLK24}, leverage Transformer-based architectures to predict and execute joint human-robot actions informed by past human behaviors and strategic goals.
This approach not only enhances the robot's capability to interpret and control interactions but also improves collaboration dynamics.

This method bolsters the robot's ability to understand and predict human strategies, a critical aspect in collaborative environments where anticipating human actions can streamline interactions and optimize task execution.
The practical applications of these models are vividly demonstrated in their capability to adapt actions in real-time based on human inputs and achieve mutual adaptation during tasks, showcasing their effectiveness in dynamic settings.

Looking forward, refining these diffusion models to more accurately translate complex and often ambiguous human instructions into precise robotic actions is crucial.
Enhancing the natural language processing capabilities within these models will allow for a more nuanced understanding and response to verbal commands.
Additionally, adopting interactive learning paradigms, where robots adjust their strategies based on real-time human feedback, could advance the development of intuitive and responsive robotic systems.

\section{Conclusion} \label{sec:conclusion}
This paper systematically reviews the application of diffusion models in planning, underscoring their potential and versatility across various domains.
First, we explored various datasets used for evaluating diffusion-based planning, ranging from motion planning and robotic tasks to reinforcement learning and multi-agent games.
Second, we summarized the integration of diffusion models in areas such as model-based and offline reinforcement learning, motion and path planning, and hierarchical planning has been shown to enhance efficiency, flexibility, and performance, addressing complex decision-making challenges.
Third, we outline the recent advancements that focus on skill learning, task planning, and trajectory generation using conditional generative approaches, which have further improved the generalization capabilities of robotic systems.
Fourth, we discuss robust safety mechanisms and uncertainty management techniques that have been integrated into diffusion-based planning models, improving their reliability in safety-critical applications.
Fifth, we find that diffusion models have also been successfully applied in specified domains such as 3D planning, demonstrating their versatility and effectiveness.
Finally, based on the above survey and discussion, we further discuss the research challenges, including improving computational efficiency and integrating diverse data sources.

\bibliographystyle{IEEEtran}
\bibliography{main}

\end{document}